\documentclass[preprint,12pt]{elsarticle}

\usepackage{amssymb}
\usepackage{amsmath}
\usepackage{graphicx}
\usepackage{booktabs}

\usepackage{float}

\journal{Advanced Engineering Informatics}

\usepackage{etoolbox}
\makeatletter
\patchcmd{\pprintMaketitle}{\hrule}{}{}{}
\patchcmd{\pprintMaketitle}{\hrule}{}{}{}
\makeatother

\begin{document}

\begin{frontmatter}


\title{YOTOnet: Zero-Shot Cross-Domain Fault Diagnosis via Domain-Conditioned Mixture of Experts}

\author[1,2]{Zesen Wang} 
\author[1]{Zihao Wu}

\author[1]{Yue Hu\corref{cor1}}
\ead{huyue1990@ecust.edu.cn} 

\author[1]{Yang Gao}

\author[1]{Fuzhen Xuan\corref{cor1}}
\ead{fzxuan@ecust.edu.cn} 

\address[1]{School of Mechanical and Power Engineering, East China University of Science and Technology, Shanghai, China}
\address[2]{Department of Mechanical Engineering and Applied Mechanics, University of Pennsylvania, Philadelphia, PA, USA}

\cortext[cor1]{Corresponding authors.}

\begin{abstract}
Mechanical equipment forms the critical backbone of modern industrial production, yet domain shift severely limits the generalization of deep learning-based fault diagnosis models across different equipment and operating conditions. Inspired by the success of foundation models in achieving zero-shot generalization, we propose YOTOnet (You Only Train Once), a novel architecture specifically designed for cross-domain fault diagnosis in mechanical equipment. YOTOnet comprises three core components: (1) a physics-aware Invariant Feature Distiller that extracts domain-agnostic representations using multi-scale dilated convolutions and FFT-based time-frequency fusion, (2) Domain-Conditioned Sparse Experts (DC-MoE) that adaptively route inputs to specialized processors via learned gating without external metadata, and (3) a dual-head classification system with auxiliary supervision. Extensive validation on five public bearing datasets (CWRU, MFPT, XJTU, OTTAWA, HUST) through 30 cross-dataset protocols demonstrates the superiority of YOTOnet compared with other state-of-the-art methods. Critically, we observe a clear scaling effect—average test F1 improves from 0.5339 (1 training dataset) to 0.705 (4 datasets), with a clear gain when moving from 3 to 4 datasets. These findings provide empirical evidence that foundation model principles can enable robust, train-once deployment for industrial fault diagnosis.
\end{abstract}


\end{frontmatter}


\section{Introduction}
\label{sec:intro}

Mechanical equipment forms the critical backbone of modern industry, with applications spanning manufacturing, transportation, and energy generation. However, unexpected failures can lead to catastrophic consequences, including production downtime, economic losses, and safety hazards \cite{Equipmentfailures}. Intelligent fault diagnosis, has emerged as a cornerstone of predictive maintenance strategies.

AI has revolutionized fault diagnosis, for it allows a new way to understand hierarchical representations from raw sensor data, which is hard for humans to understand at first sight, eliminating the need for manual feature engineering \cite{s23031305}. However, a new challenge appears: domain shift. Mostly, models perform well on their training datasets, but have poor performance on unseen datasets. This defect makes it hard for industrial application in diagnosis of machinery.

Existing approaches to address domain shift fall into two main categories: domain adaptation and domain generalization \cite{wang2022generalizing}. In recent domain adaptation research, Jiang utilized a mechanism-driven adversarial network to achieve interpretable fault diagnosis across different operating conditions \cite{JIANG2025112244}, while Pan developed a self-supervised contrastive network to mitigate class boundary confusion and obtain superior performance \cite{PAN2025103058}. In the domain generalization field, Yang proposed a mixed domain fusion network to capture invariant features for generalizing to unknown real-world conditions \cite{Yang10955333}, and Cui employed a causal decoupling network to accurately model the data generation process, achieving high stability for multi-bearing fault diagnosis \cite{CUI2025236}.

The great generalization capabilities of Large Language Models (LLMs) have inspired new methods in different fields. The paradigm of large models has been successfully applied across diverse scientific fields. Meta AI introduced the Segment Anything Model \cite{kirillov2023segment}, a powerful vision foundation model that achieves remarkable zero-shot segmentation. In chemistry, researchers presented ChemCrow \cite{bran2023chemcrowaugmentinglargelanguagemodels}, an LLM augmented with tools that can autonomously plan complex chemical reactions. Furthermore, Salesforce AI's Nucleotide Transformer \cite{DallaTorre2023} applies large-scale pre-training to DNA sequences, effectively learning the language of life to predict genetic functions. Consequently, the application of LLM methodologies to fault diagnosis is becoming increasingly prevalent. FD-LLM \cite{qaid2024fdllmlargelanguagemodel} adapts Large Language Models by textualizing vibration signals. Rm-GPT \cite{wang2025rmgptfoundationmodelgenerative} was developed as a foundation model tailored specifically for mechanical equipment data. However, as a recent survey \cite{Chen_Lei_Li_Parkinson_Li_Liu_Lu_Wang_Wang_Yang_Ye_Zhao_2025} highlights, significant challenges remain, emphasizing the need for true zero-shot generalization.

This paper presents YOTO, a foundation specifically designed for zero-shot cross-domain fault diagnosis in mechanical equipment. The evaluation across five public datasets reveals a significant scaling effect: average F1 improves from 0.5339 to 0.705 as the training source expands from one to four domains. While designed for zero-shot deployment, YOTO also supports rapid few-shot adaptation, achieving 0.99 F1 with 256 labeled samples via Q-LoRA.

The main contributions of this work are:
\begin{itemize}
    \item We propose the ``You Only Train Once'' (YOTO) paradigm for industrial fault diagnosis. Unlike traditional domain adaptation, YOTO builds a generalist foundation model from multiple source domains that generalizes to unseen operating conditions and equipment without target data or fine-tuning.

    \item We design a physics-aware Invariant Feature Distiller that combines multi-scale dilated convolutions with FFT-based time-frequency fusion and dual attention mechanisms. This module extracts robust, domain-invariant ``physical tokens'' from raw vibration signals to mitigate spectral shifts.

    \item We introduce a Domain-Conditioned Sparse Mixture-of-Experts (DC-MoE) where routing is conditioned purely on the input signal (no external domain IDs). This allows dynamic selection of specialized experts for varying signal characteristics while ensuring diverse utilization via load balancing.

    \item We provide systematic empirical evidence of scaling laws in cross-domain fault diagnosis. Evaluations on 30 train/test splits show that increasing training domain diversity (from 1 to 4) yields consistent zero-shot performance gains, validating the path toward industrial foundation models.
\end{itemize}

The remainder of this paper is organized as follows. Section 2 details the YOTO methodology, including architecture design and training strategy. Section 3 presents comprehensive experimental results, including scaling law analysis and ablation studies. Section 4 concludes with implications and future directions.

\section{YOTOnet Network Architecture}
\label{sec:methodology}

\subsection{Problem Setup and Design Motivations}

We consider $K$ source domains $\mathcal{S}=\{D_1,\dots,D_K\}$ and an unseen target domain $D_t$. Each domain provides samples $(signals,label)$. Under zero-shot domain generalization, we learn parameters $\theta$ without seeing $D_t$ by minimizing the expected supervised risk on the sources:
\begin{equation}
\min_{\theta} \; \mathbb{E}_{s\sim \mathcal{S}}\,\mathbb{E}_{(x,y)\sim D_s} \, \ell\big(f_{\theta}(x), y\big), \quad\text{and evaluate on } D_t \text{ without adaptation.}
\end{equation}

Empirically, cross-domain shifts arise from (i) spectral-shape differences and operating conditions; (ii) multi-scale temporal cues; and (iii) domain-specific processing preferences. Accordingly, YOTOnet is designed with three principles that align with these factors:
\begin{itemize}
    \item Physics-aware time-frequency tokenization and invariant feature distillation to mitigate spectral-shape and operating-condition shifts;
    \item Multi-scale temporal modeling (e.g., dilated CNN branches with residual links) to capture transient, multi-resolution fault signatures;
    \item Domain-conditioned sparse expert routing to accommodate domain-specific processing preferences while preserving domain-agnostic fault semantics;
\end{itemize}

\subsection{Method Overview}

\begin{figure}[H]
    \centering
    \includegraphics[width=\textwidth]{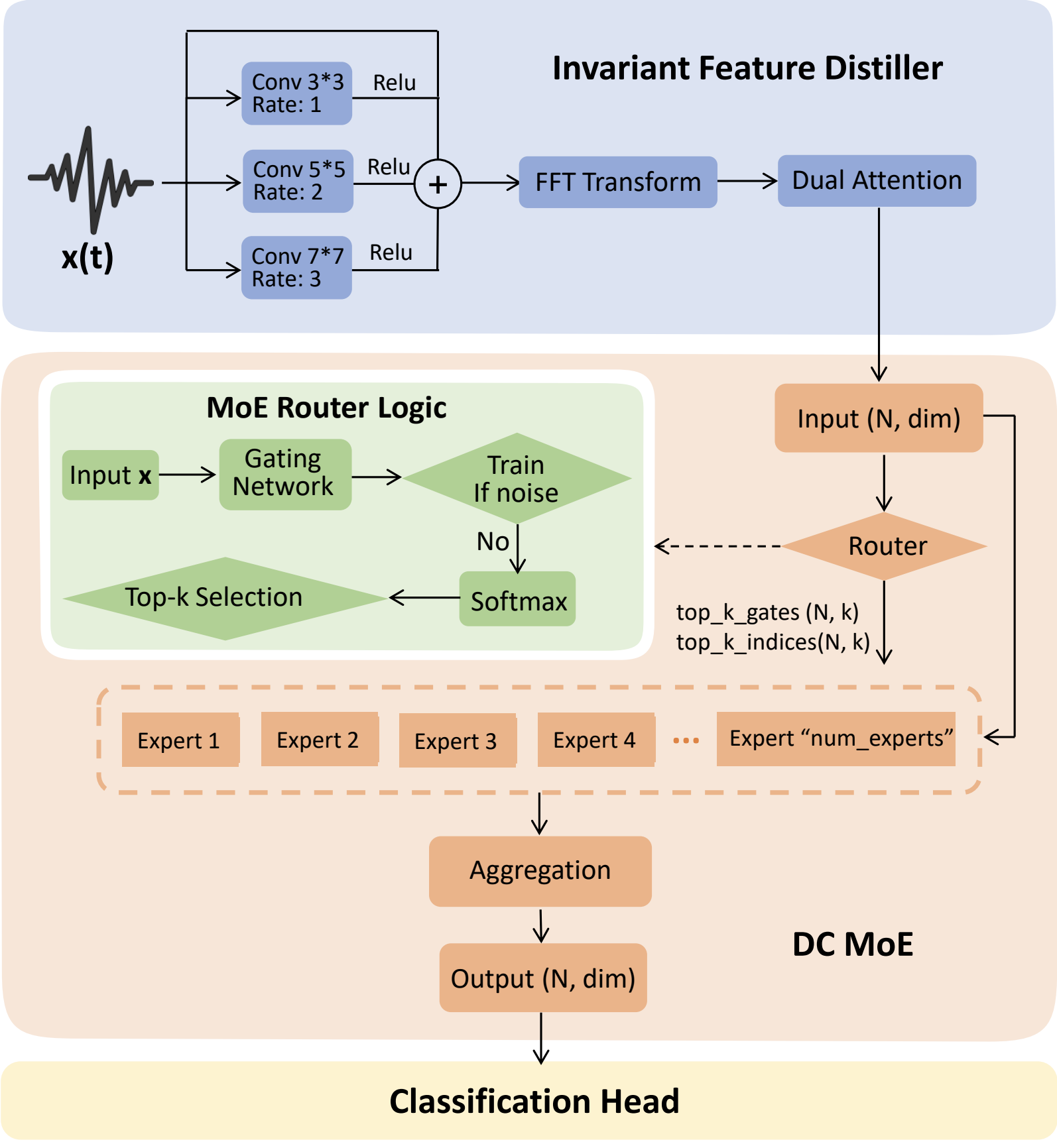}
    \caption{YOTOnet model architecture. The model consists of three core components: Invariant Feature Distiller for extracting domain-invariant features, Domain-Conditioned Sparse Experts (DC\textendash MoE) for adaptive processing, and Classification Head for fault prediction.}
    \label{fig:architecture}
\end{figure}

We target YOTO by constructing a modular foundation model that processes raw signals into a shared 'fault dictionary' without domain-specific adaptation. The overall architecture of YOTOnet is illustrated in Figure~\ref{fig:architecture}. It comprises three core components: an Invariant Feature Distiller, Domain-Conditioned Sparse Experts (DC\textendash MoE), and a Classification Head.

Subsequent experiments applying this method demonstrate that significant scaling effects also exist in the industrial sensor data domain. This means that by increasing training data volume and enhancing model complexity, we can observe simultaneous significant improvements in model diagnostic performance and cross-domain generalization capability, ultimately manifesting as powerful multi-domain generalization ability.

\subsection{Invariant Feature Distiller}

\begin{figure}[H]
    \centering
    \includegraphics[width=1.0\textwidth]{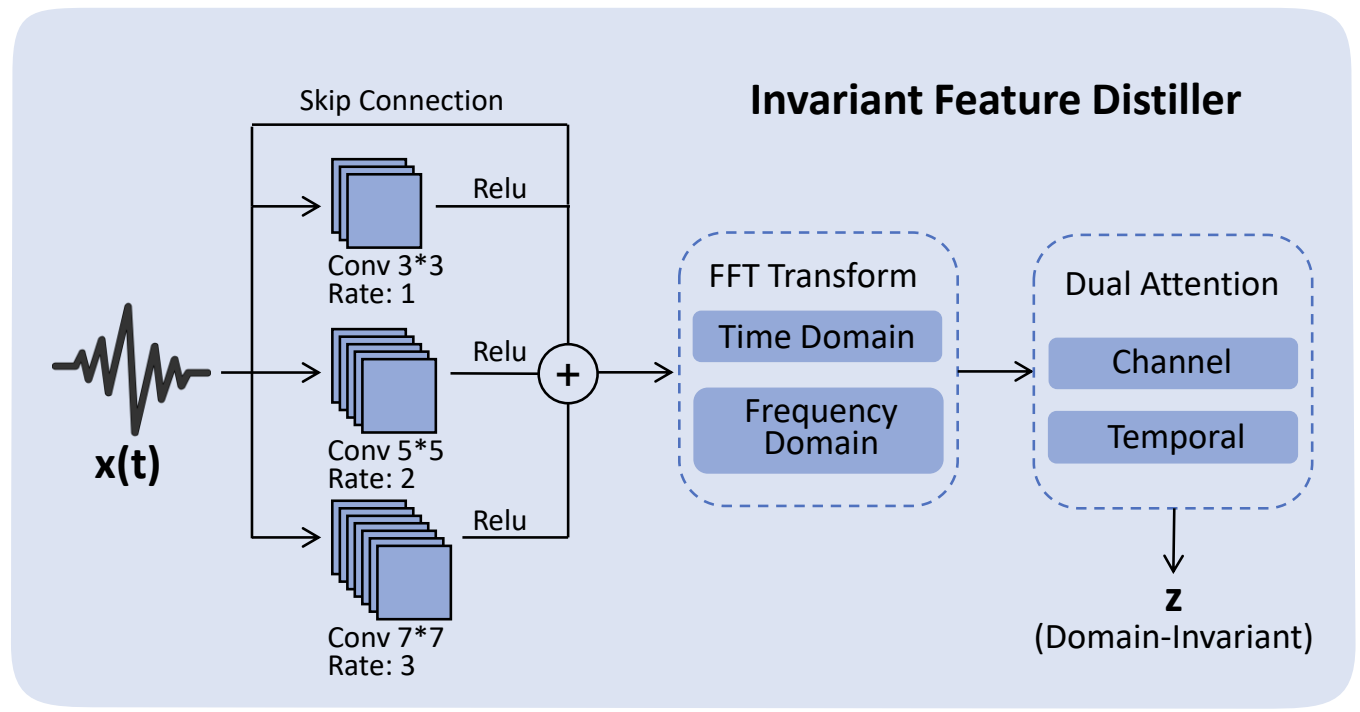}
    \caption{Detailed structure of the Invariant Feature Distiller. The module employs multi-scale convolutional branches, dilated convolutions, residual connections, FFT feature fusion, and dual attention mechanisms.}
    \label{fig:distiller}
\end{figure}

As the front-end encoder, the Invariant Feature Distiller forms domain-invariant "physical tokens" using a compact multi-scale CNN \cite{szegedy2015going} with dilated convolutions \cite{yu2016multi} and residual links \cite{he2016deep}. As detailed in Figure~\ref{fig:distiller}, the distiller utilizes a parallel three-branch structure with varying dilation rates ($3 \times 3, 5 \times 5, 7 \times 7$) to capture multi-resolution temporal signatures.

We enrich time-frequency cues via FFT-based fusion of frequency-domain and time-domain features, which is a common apply in vibration analysis. And applying dual attention \cite{hu2018squeeze} to emphasize informative channels.

\subsection{Domain-Conditioned Sparse Experts (DC\textendash MoE)}

\begin{figure}[!htbp]
    \centering
    \includegraphics[width=0.8\textwidth]{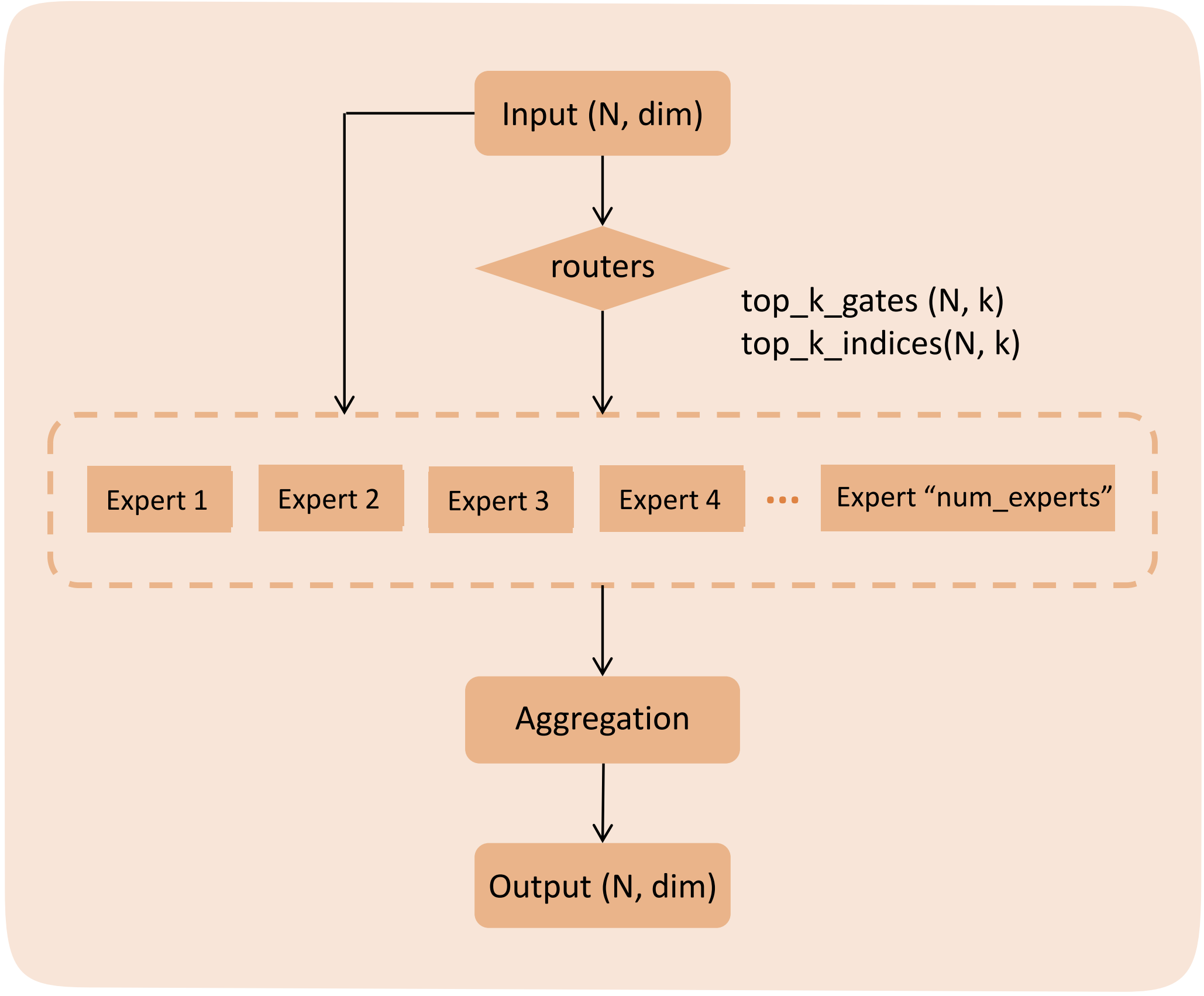}
    \caption{Overall architecture of Domain-Conditioned Sparse Experts (DC\textendash MoE). The system routes input features through a gating network to selectively activate top-k experts, combining their outputs via weighted aggregation. In our implementation, experts are feed-forward networks.}
    \label{fig:dcmoe}
\end{figure}

We extend the router to be \emph{domain-conditioned via the signal itself} (no external metadata), as illustrated in Figure~\ref{fig:dcmoe}. Let $z$ denote features from the distiller. The gate produces logits and sparse probabilities
\begin{equation}
\mathbf{p} = \mathrm{softmax}(W_g z), \quad \mathbf{m}=\text{Top-}k(\mathbf{p}), \quad y=\sum_{i} m_i \, p_i \, h_i(z),
\end{equation}
where $h_i$ is the $i$-th expert and $\mathbf{m}$ masks all but the $k$ largest entries (straight-through during backprop).

\subsubsection{Expert Architecture (FFN-based DC\textendash MoE)}

As highlighted in Figure~\ref{fig:dcmoe}, each expert $h_i(\cdot)$ is a lightweight feed-forward network (FFN) that maps the distilled feature vector $z \in \mathbb{R}^d$ to an enhanced representation $h_i(z)$. Concretely, $h_i(z)=W_{2,i}\,\phi(W_{1,i} z + b_{1,i}) + b_{2,i}$, where $\phi$ is a ReLU activation. When sequence features $Z \in \mathbb{R}^{T\times d}$ are used, the FFN is applied token-wise with shared weights and followed by mean/attention pooling to form the expert output. This follows common MoE practice and keeps compute predictable and efficient.

\subsubsection{Gating Network and Routing Logic}

\begin{figure}[!htbp]
    \centering
    \includegraphics[width=1.0\textwidth]{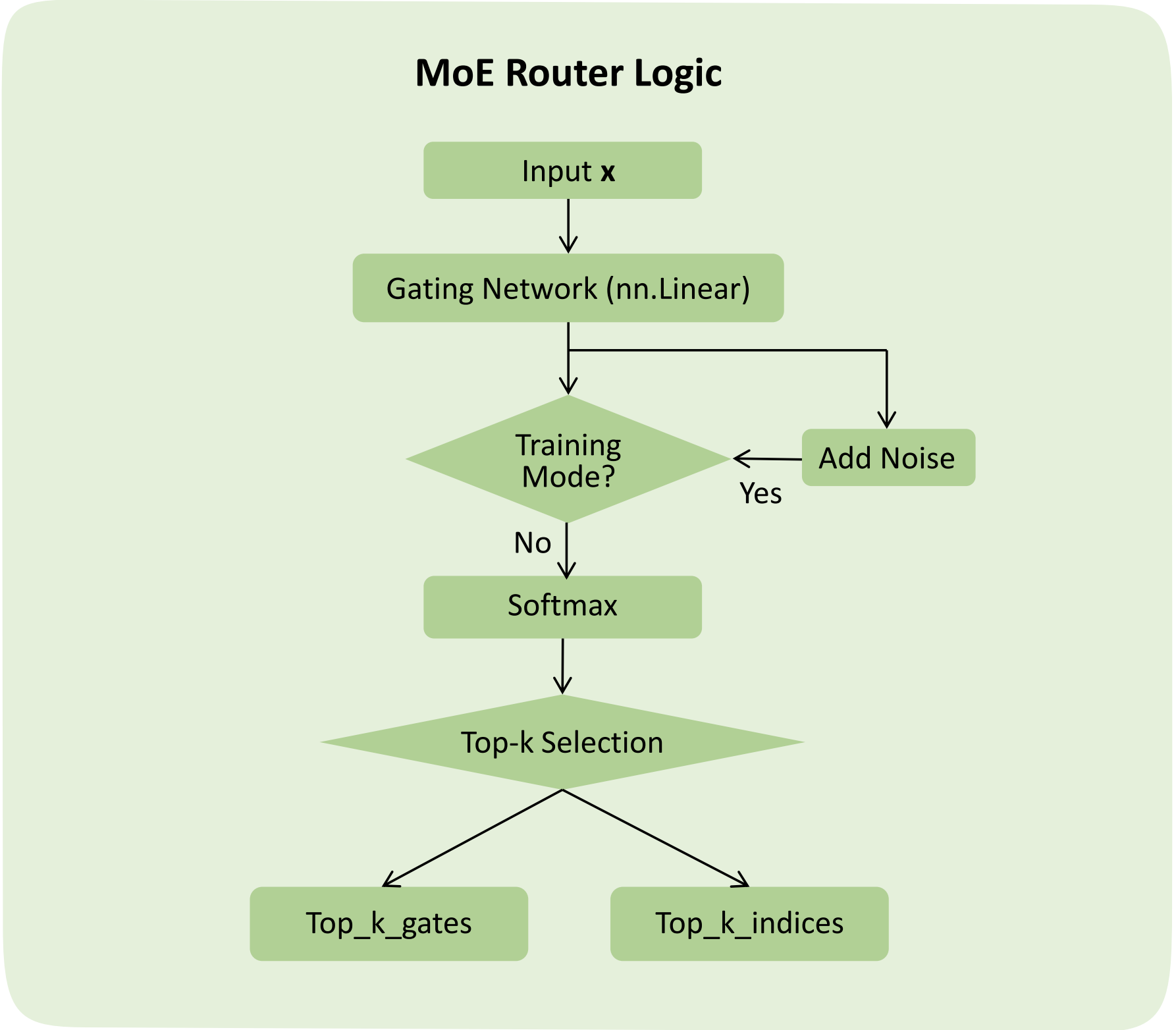}
    \caption{Gating network routing logic. The gate network transforms input features $z$ into expert selection probabilities $\mathbf{p}$ via $W_g$ and softmax. The Top-$k$ operation produces a binary mask $\mathbf{m}$ that activates only the most relevant experts, with straight-through gradient estimation during backpropagation.}
    \label{fig:gating}
\end{figure}

The gating mechanism determines which experts process each input, as detailed in Figure~\ref{fig:gating}. Given distilled features $z$, the gate network $W_g$ computes expert affinities, applies softmax normalization to obtain probabilities $\mathbf{p}$, and selects the top-$k$ experts via a binary mask $\mathbf{m}$. During forward pass, only the selected experts are activated; during backpropagation, we use straight-through estimators to approximate gradients through the discrete Top-$k$ operation. This sparse activation reduces computational cost while maintaining model capacity.

\subsubsection{Load Balancing Regularization}

\begin{figure}[!htbp]
    \centering
    \includegraphics[width=1.0\textwidth]{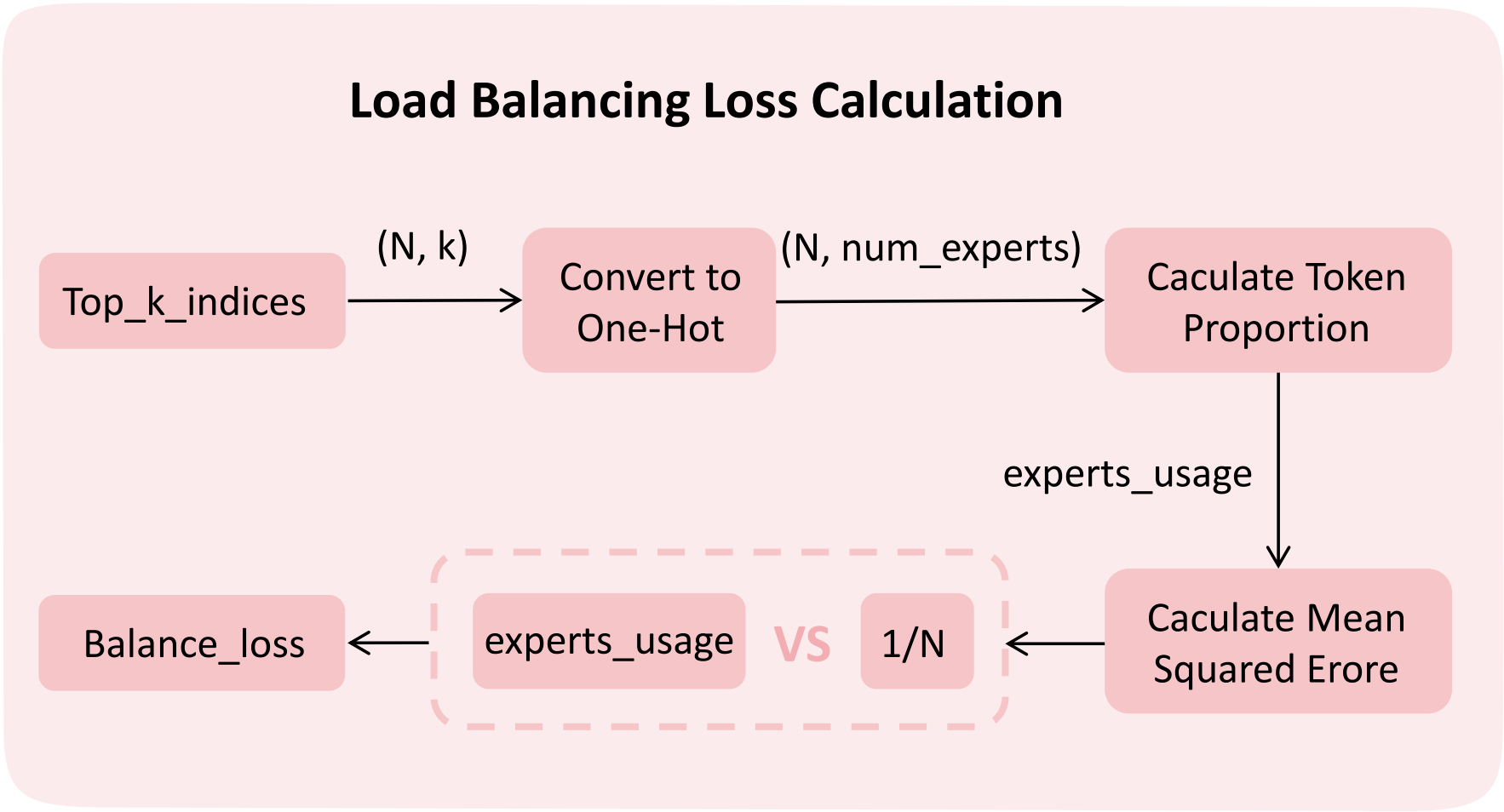}
    \caption{Load balancing regularizer computation. The regularizer measures the squared deviation between actual expert utilization fractions $f_i$ and the uniform distribution $1/N$. This prevents expert collapse by penalizing scenarios where certain experts are over-utilized while others remain idle.}
    \label{fig:load_balancing}
\end{figure}

To prevent expert collapse (where only a few experts are consistently selected) and promote balanced utilization, we add a load balancing regularizer as shown in Figure~\ref{fig:load_balancing}:
\begin{equation}
\label{eq:gate}
\mathcal{L}_{\text{gate}}= \lambda_{\text{bal}} \sum_i \left( f_i - \frac{1}{N} \right)^2,
\end{equation}
where $f_i$ is the fraction of samples routed to expert $i$ and $N$ is the total number of experts. This MSE-based regularizer penalizes deviations from uniform expert utilization across the batch, encouraging the model to distribute workload evenly among all experts. The hyperparameter $\lambda_{\text{bal}}$ controls the strength of this regularization.

\subsubsection{Experts Network}

The experts in DC\textendash MoE do not introduce additional recurrence or
attention; each expert $h_i(\cdot)$ is implemented as a compact FFN. Given a distilled feature vector $z$, the expert applies a
linear projection, a point-wise ReLU nonlinearity, and a final
linear projection to produce $h_i(z)$. For sequence features, the same FFN is
applied token-wise with shared weights, and the resulting token representations
are aggregated by pooling. This simple FFN design keeps each expert lightweight
and efficient while still allowing the overall MoE layer to capture diverse
processing patterns across domains.

\subsection{Classification Head}

A main classifier aggregates sequence features via attention pooling followed by an MLP and Softmax to predict fault categories. An auxiliary head, attached earlier in the network, provides additional supervision to improve feature quality and stabilize training.

\subsection{Training Objective and Loss Function}

We minimize a weighted sum consisting of: (i) a cross-entropy loss for the main classifier, (ii) an auxiliary cross-entropy for the early head, and (iii) the load balancing regularizer $\mathcal{L}_{\text{gate}}$ in Eq.~(\ref{eq:gate}) for DC\textendash MoE.

The total objective is
\begin{equation}
\mathcal{L}_{\text{total}} = \mathcal{L}_{\text{main}} + \alpha \, \mathcal{L}_{\text{aux}} + \beta \, \mathcal{L}_{\text{gate}},
\end{equation}
where $\alpha$ and $\beta$ weight the auxiliary and load balancing terms, respectively.
\subsection{Optional Few-shot Adaptation with Q-LoRA}

Although our core setting is zero-shot, we also consider a practical few-shot variant for rapid on-site adaptation. We adopt Q-LoRA, which keeps the backbone quantized (e.g., 4-bit) and trains low-rank adapters on a small labeled set. In our experiments, fine-tuning with only 256 labeled samples yields an average F1 of 0.99 on the same test protocol. This result is reported separately and does not alter the zero-shot conclusions above.

\section{Experimental Validation}
\label{sec:experiments}

\setlength{\tabcolsep}{1.8pt}
\renewcommand{\arraystretch}{0.78}

\subsection{Data Processing and Preparation}

We jointly train on five public bearing datasets—CWRU \cite{li2019cwru}, MFPT, XJTU \cite{wang2018hybrid}, OTTAWA \cite{sehri2023university}, and HUST \cite{LIU201833}—covering diverse operating conditions and fault modes. For all experiments we construct a binary label space with two fault types (inner-race vs. outer-race fault), and all signals are (re)sampled to 25.6 kHz for consistency.

Table~\ref{tab:dataset_stats} summarizes the basic statistics of the five datasets used in this study. For each benchmark, we list the total number of samples, the number of inner- and outer-race fault samples, and the unified sampling frequency. These benchmarks span a variety of operating conditions such as different motor speeds, loads, and fault severities; we follow the standard working conditions and segmenting protocols defined in the original publications.

\begin{table}[h!]
    \centering
    \caption{Statistics of the five bearing datasets used in this study.}
    \label{tab:dataset_stats}
    \begin{tabular}{lcccc}
        \toprule
        \textbf{Dataset} & \textbf{Total samples} & \textbf{Inner fault} & \textbf{Outer fault} & \textbf{Sampling frequency} \\
        \midrule
        CWRU & 6712 & 2149 & 4563 & 25.6 kHz \\
        MFPT & 363 & 126 & 237 & 25.6 kHz \\
        XJTU & 17040 & 12528 & 4512 & 25.6 kHz \\
        HUST & 1860 & 930 & 930 & 25.6 kHz \\
        OTTAWA & 1240 & 620 & 620 & 25.6 kHz \\
        \bottomrule
    \end{tabular}
\end{table}

\subsection{Experimental Setup and Environment}

To systematically evaluate the model's generalization capability under different data combinations and the impact of data scale on performance, we designed a series of cross-dataset ablation experiments. The experimental setup is as follows: we select training and test sets from all five datasets, with the key constraint that all five datasets must be used and assigned to either the training or test set. By traversing all possibilities where the training set contains 1 to 4 datasets (with the remaining datasets forming the test set), we generate a total of $C_5^1 + C_5^2 + C_5^3 + C_5^4 = 30$ different training/test set partition combinations. Hyperparameter settings remain unchanged across all experiments. Similarly, training-related hyperparameters such as training epochs, learning rate, and optimizer also remain consistent across all 30 experiments. The primary evaluation metric is the sample-averaged F1 score on the test set, while also recording the F1 score for each category. All experiments were conducted in the following hardware and software environment:

\begin{table}[h!]
\centering
\caption{Experimental Environment Configuration}
\label{tab:exp_env}
\begin{tabular}{ll}
\toprule
\textbf{Item} & \textbf{Specification} \\
\midrule
CPU & 14 vCPU Intel(R) Xeon(R) Gold 6348 @ 2.60GHz \\
GPU & 1x NVIDIA A800-80GB \\
RAM & 100GB \\
Operating System & Ubuntu 22.04 \\
Python Version & Python 3.10 \\
CUDA Version & CUDA 11.8 \\
Package Manager & Miniconda3 \\
Disk Space & System: 30GB, Data: 100GB \\
\bottomrule
\end{tabular}
\end{table}

\subsection{Method Comparison}

\begin{figure}[H]
    \centering
    \includegraphics[width=1.0\textwidth]{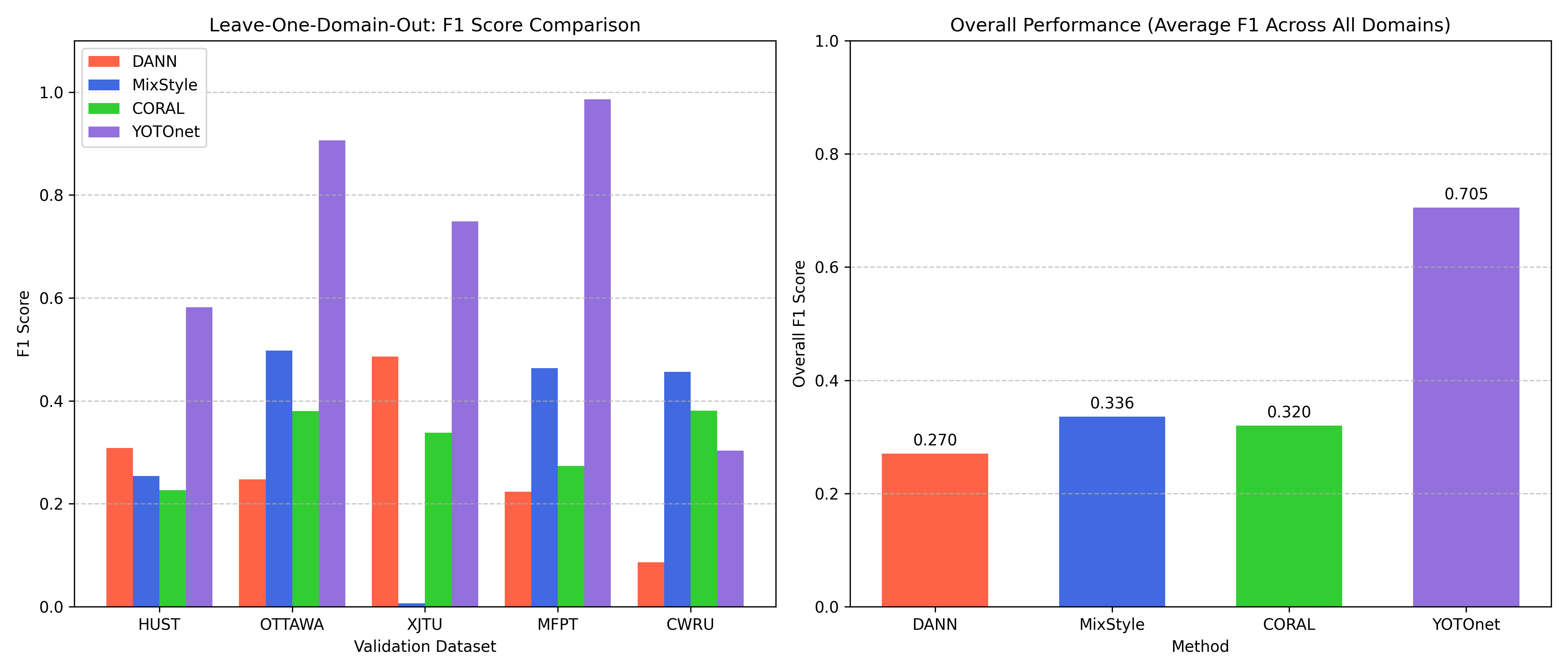}
    \caption{Bar-chart visualization of method performance across the five test sets.}
    \label{fig:comparison_bar}
\end{figure}

We compare YOTOnet with DANN \cite{ganin2016domainadversarialtrainingneuralnetworks}, CORAL \cite{sun2016deepcoralcorrelationalignment}, and MixStyle \cite{zhou2021domaingeneralizationmixstyle}, and the comparison results are given in Table~\ref{tab:method_comparison} and Figure~\ref{fig:comparison_bar}. Table~\ref{tab:method_comparison} reports average F1 on each test set.YOTOnet achieves the best performance on most of all datasets except CWRU and it reaches 0.9862 (F1 score) on MFPT.

\begin{table}[t]
\centering
\caption{Performance of Different Models on Various Test Sets}
\label{tab:method_comparison}
\begin{tabular}{lccccc}
\toprule
\textbf{Test Set/Model} & \textbf{HUST} & \textbf{OTTAWA} & \textbf{XJTU} & \textbf{MFPT} & \textbf{CWRU} \\
\midrule
YOTOnet & \textbf{0.5821} & \textbf{0.9065} & \textbf{0.7485} & \textbf{0.9862} & 0.3033 \\
CORAL & 0.2266 & 0.3803 & 0.3382 & 0.2737 & 0.3812 \\
DANN & 0.3082 & 0.2477 & 0.4861 & 0.2239 & 0.0863 \\
MixStyle & 0.2538 & 0.4980 & 0.0069 & 0.4638 & \textbf{0.4568} \\
\bottomrule
\end{tabular}
\end{table}

\vspace{2em}

\subsection{Ablation Studies}

To understand the contribution of each component in YOTOnet, we further conduct ablation studies on six model variants: (1) Full, the complete model with domain-conditioned sparse experts, FFT-based feature fusion, dual attention, and gate load-balancing regularization; (2) Random Expert, replacing the learned gate with random expert selection while keeping the experts unchanged, to evaluate the importance of domain-conditioned routing; (3) No Balance, removing the gate load-balancing loss from the training objective, to assess the effect of expert load balancing; (4) Avg Fusion, averaging the outputs of all experts with equal weights instead of sparse top-$k$ routing, to test the benefit of sparse MoE fusion; (5) No FFT, removing the frequency-domain FFT branch in the feature extractor and using only time-domain features, to verify the contribution of spectral information; (6) No Dual Attn, removing the dual attention or squeeze-and-excitation modules in the feature extractor, to assess the impact of attention-based feature reweighting.

Table~\ref{tab:ablation_all} reports the F1 score of these variants when each of the five datasets is treated as the unseen target domain. For each dataset we also indicate the best-performing variant and the gap between the full model and the best variant.

\vspace{1\baselineskip} 

\begin{table}[h!]
    \centering
    \caption{Ablation studies on five test domains. We report F1 score of the full model and five ablated variants; 'Best model' denotes the highest-scoring variant per dataset, and 'Full vs. Best' summarizes the performance gap between the full model and the best variant.}
    \label{tab:ablation_all}
    \scriptsize
    \begin{tabular}{lccccccccc}
        \toprule
        \textbf{Dataset} & \textbf{Full} & \shortstack[c]{\textbf{Random}\\\textbf{Expert}} & \shortstack[c]{\textbf{No}\\\textbf{Balance}} & \shortstack[c]{\textbf{Avg}\\\textbf{Fusion}} & \shortstack[c]{\textbf{No}\\\textbf{FFT}} & \shortstack[c]{\textbf{No}\\\textbf{Dual}\\\textbf{Attn}} & \shortstack[c]{\textbf{Best}\\\textbf{model}} & \shortstack[c]{\textbf{Best}\\\textbf{F1}} & \shortstack[c]{\textbf{Full vs.}\\\textbf{Best}} \\
        \midrule
        XJTU   & 0.7428 & 0.3695 & 0.2648 & 0.6467 & 0.5877 & 0.6353 & Full          & 0.7428 & Best \\
        MFPT   & 0.9862 & 0.5785 & 0.3471 & 0.3471 & 0.7631 & 0.7107 & Full          & 0.9862 & Best \\
        CWRU   & 0.3582 & 0.3258 & 0.3202 & 0.3202 & 0.4242 & 0.4513 & No Dual Attn  & 0.4513 & $-0.0931$ \\
        HUST   & 0.6371 & 0.6661 & 0.6817 & 0.5 & 0.671 & 0.5806 & No Balance    & 0.6817 & $-0.0446$ \\
        OTTAWA & 0.9065 & 0.9952 & 0.5 & 0.9185 & 1 & 0.6282 & No FFT       & 1 & $-0.0935$ \\
        \bottomrule
    \end{tabular}
\end{table}

\vspace{1\baselineskip} 

As shown in Table~\ref{tab:dataset_stats}, XJTU is the largest and most imbalanced
test domain (17{,}040 segments with an inner/outer ratio of roughly
2.8:1). In this more challenging regime, the full YOTOnet consistently achieves
the best F1 (0.7428), while all ablated variants degrade noticeably. This
indicates that domain-conditioned sparse experts, FFT-based fusion, dual
attention, and gate load-balancing work together to exploit diverse operating
conditions without collapsing onto the majority class.

By contrast, the behavior on HUST and OTTAWA is more mixed: some ablated
variants slightly outperform the full model. Both benchmarks are small (1{,}860
and 1{,}240 samples) and perfectly balanced (inner:outer = 1:1), which reduces
the benefit of the gate-balancing regularizer and makes results more sensitive
to sampling noise. In addition, these domains appear closer to some of the
training domains in terms of operating conditions, so simpler variants (e.g.,
without balance regularization or FFT) can already fit them well. We therefore
view these gains as benign variance rather than evidence against the full
design, and retain the full model as the default since it provides the most
robust performance across the larger and more challenging domains.

On CWRU, all methods achieve comparatively low F1 (around 0.3--0.4 in Table~\ref{tab:method_comparison}), indicating a substantial domain gap between
CWRU and the other datasets. This challenging case underscores the inherent
difficulty of zero-shot cross-dataset diagnosis when operating conditions
differ drastically.

\begin{figure}[H]
    \centering
    \includegraphics[width=1.0\textwidth]{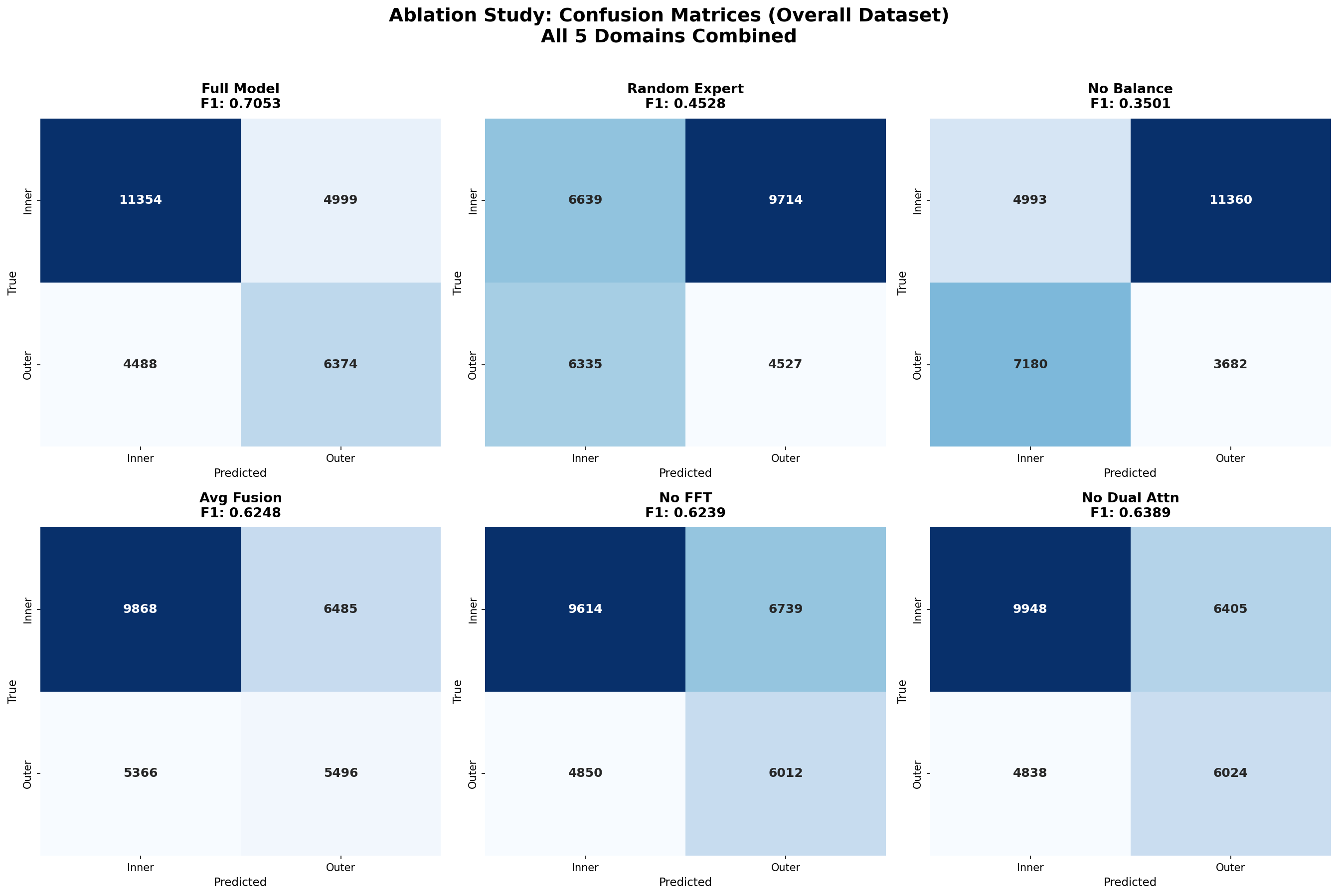}
    \caption{Combined confusion matrices for representative zero-shot evaluations across test domains.}
    \label{fig:confusion_matrices}
\end{figure}

\subsection{Validation of Scaling Effects}

We quantify scaling by evaluating 30 train/test splits formed by selecting 1--4 datasets for training from {CWRU, MFPT, XJTU, OTTAWA, HUST} and validating on the held-out datasets. Table~\ref{tab:tasks_all} lists all training/test configurations for Tasks 1--4. And Table~\ref{tab:scaling_results} summarizes the average performance—F1 score—obtained by the model on the test set when the training set consists of different numbers of original datasets in these experiments.

As Table~\ref{tab:scaling_results} shows, increasing training datasets from 1 to 4 steadily boosts average test F1, with a clear improvement from 0.6409 to 0.705 when moving from 3 to 4 datasets, confirming the scaling effect on industrial sensor data.

\subsection{Analysis of Experimental Results}

In summary, YOTOnet (i) consistently outperforms DANN, CORAL, and MixStyle across all test sets; (ii) exhibits clear scaling--average F1 rises from 0.5339 (1 dataset) to 0.705 (4 datasets), with consistent gains as more training domains are included; (iii) achieves strong zero-shot generalization (e.g., 0.9862 on the MFPT dataset), indicating domain-agnostic feature learning; and (iv) benefits from its three-module design for robust end-to-end generalization. Collectively, these results substantiate the YOTO paradigm and provide empirical evidence of scaling on industrial sensor data.

\vspace{1\baselineskip} 

\begin{table}[!h]
\centering
\caption{Training configurations for scaling experiments (Tasks 1--4). Task 1 uses a single training dataset, while Tasks 2--4 use two, three, and four datasets, respectively.}
\label{tab:tasks_all}
\scriptsize
\begin{tabular}{lll}
\toprule
\textbf{Task} & \textbf{Training Set} & \textbf{Test Set} \\
\midrule
Task 1 & CWRU   & XJTU OTTAWA MFPT HUST \\
Task 1 & XJTU   & CWRU OTTAWA MFPT HUST \\
Task 1 & OTTAWA & CWRU XJTU MFPT HUST \\
Task 1 & MFPT   & XJTU OTTAWA CWRU HUST \\
Task 1 & HUST   & CWRU XJTU OTTAWA MFPT \\
\midrule
Task 2 & CWRU MFPT   & XJTU OTTAWA HUST \\
Task 2 & CWRU XJTU   & OTTAWA MFPT HUST \\
Task 2 & CWRU OTTAWA & HUST XJTU MFPT \\
Task 2 & CWRU HUST   & MFPT OTTAWA XJTU \\
Task 2 & MFPT XJTU   & HUST OTTAWA CWRU \\
Task 2 & MFPT HUST   & CWRU OTTAWA XJTU \\
Task 2 & MFPT OTTAWA & CWRU HUST XJTU \\
Task 2 & XJTU OTTAWA & CWRU HUST MFPT \\
Task 2 & XJTU HUST   & CWRU OTTAWA MFPT \\
Task 2 & OTTAWA HUST & CWRU XJTU MFPT \\
\midrule
Task 3 & CWRU MFPT XJTU     & HUST OTTAWA \\
Task 3 & CWRU MFPT OTTAWA  & HUST XJTU \\
Task 3 & CWRU MFPT HUST    & OTTAWA XJTU \\
Task 3 & CWRU XJTU OTTAWA  & HUST MFPT \\
Task 3 & CWRU XJTU HUST    & OTTAWA MFPT \\
Task 3 & CWRU OTTAWA HUST  & XJTU MFPT \\
Task 3 & MFPT XJTU OTTAWA  & HUST CWRU \\
Task 3 & MFPT XJTU HUST    & OTTAWA CWRU \\
Task 3 & MFPT OTTAWA HUST  & CWRU XJTU \\
Task 3 & XJTU OTTAWA HUST  & CWRU MFPT \\
\midrule
Task 4 & CWRU MFPT XJTU OTTAWA  & HUST \\
Task 4 & CWRU MFPT XJTU HUST   & OTTAWA \\
Task 4 & CWRU MFPT OTTAWA HUST & XJTU \\
Task 4 & CWRU XJTU OTTAWA HUST & MFPT \\
Task 4 & MFPT XJTU OTTAWA HUST & CWRU \\
\bottomrule
\end{tabular}
\end{table}

\vspace{1\baselineskip} 

\begin{table}[h!]
    \centering
    \caption{Average Performance Under Different Training Set Scales}
    \label{tab:scaling_results}
    \scriptsize

    \begin{tabular}{lc}

    \toprule
    \textbf{Task} & \textbf{Average Test F1} \\
    \midrule
    Task 1 (1 dataset) & 0.5339 \\
    Task 2 (2 datasets) & 0.6135 \\
    Task 3 (3 datasets) & 0.6409 \\
    Task 4 (4 datasets) & 0.705 \\
    \bottomrule
    \end{tabular}
\end{table}

\subsection{Few-shot Fine-tuning with Q-LoRA (256 Samples)}

For completeness, we further evaluate an optional few-shot adaptation. Using Q-LoRA with 256 labeled samples, YOTOnet attains an average F1 of 0.99 on the same test splits. As summarized in Table~\ref{tab:qlora_avg}, this reuslt is reported separately and is not included in the zero-shot tables above (Table~\ref{tab:method_comparison}), since it uses additional labeled target-domain data.

\begin{table}[!h]
    \centering
    \caption{Few-shot Q-LoRA (256 samples) vs. zero-shot YOTOnet: average F1 on the same test protocol.}
    \label{tab:qlora_avg}
    \scriptsize
    \begin{tabular}{l c}
        \toprule
        \textbf{Model} & \textbf{Avg F1} \\
        \midrule
        Zero-shot YOTOnet (4-source) & 0.705 \\
        Q-LoRA (256 labeled samples) & 0.99 \\
        \bottomrule
    \end{tabular}
\end{table}

\section{Conclusion and Future Work}
\label{sec:conclusion}

We presented YOTO (You Only Train Once) and validated YOTOnet—a three-module architecture (Invariant Feature Distiller, Domain-Conditioned Sparse Experts, Classification Head)—via joint training on five public bearing datasets (CWRU, MFPT, XJTU, OTTAWA, HUST) and zero-shot evaluation. Key findings include: (i) train-once generalization to unseen domains without target data or fine-tuning; (ii) consistent empirical scaling across 30 cross-dataset splits, with average test F1 improving from 0.5339 (1 dataset) to 0.705 (4 datasets), with clear gains when moving from 3 to 4 datasets; and (iii) superiority over DANN, CORAL, and MixStyle on all test sets. We hope this proposed work serves as a foundational empirical step toward the realization of universal, trustworthy industrial foundation models.

However, the proposed method still has three limitations to be addressed in the future: (i) improving data efficiency via few-shot learning or meta-learning; (ii) integrating multi-modal signals (e.g., temperature, current, and acoustics) to enhance robustness; and (iii) strengthening interpretability and reliability.

\bibliographystyle{elsarticle-num}
\bibliography{references}

\end{document}